\documentclass{article}

\usepackage[utf8]{inputenc}
\usepackage[utf8]{inputenc}
\usepackage{graphicx}
\usepackage{amsfonts}
\usepackage{amsmath}
\usepackage{wrapfig}
\usepackage{authblk}

\usepackage{lineno,hyperref}
\modulolinenumbers[5]
\usepackage[caption=false]{subfig}

\usepackage[english]{babel}

\title{My Health Sensor, my Classifier – Adapting a Trained Classifier to Unlabeled End-User  Data}

\author[1]{\small Konstantinos Nikolaidis \thanks{Corresponding Author: konstan@ifi.uio.no}}

\author[1]{\small Stein Kristiansen}
\author[1]{\small Thomas Plagemann}
\author[1]{\small Vera Goebel}
\author[1]{\small Knut Liestøl}

\author[2]{\small Mohan Kankanhalli}
\author[3,4,5]{\small Gunn Marit Traaen}
\author[6]{\small Britt Øverland}
\author[4,10]{\small Harriet Akre}
\author[7,8]{\small Lars Aakerøy}
\author[7,8]{\small Sigurd Steinshamn}

\affil[1]{\scriptsize Department of Informatics, University of Oslo, Norway}

\affil[2]{Department of Computer Science, National University of Singapore, Singapore}

\affil[3]{Department of Cardiology, Oslo University Hospital, Rikshospitalet, Oslo, Norway}
\affil[4]{Institute of Clinical Medicine, Faculty of Medicine, University of Oslo, Oslo, Norway }
\affil[5]{KG Jebsen Center for Cardiac Research, University of Oslo, Norway and Center for Heart Failure Research, Oslo University Hospital, Norway}

\affil[6]{Department of Otorhinolaryngology, Head \& Neck Surgery, Sleep Unit, Lovisenberg Diakonale Hospital, Oslo, Norway}
\affil[7]{Department of Thoracic Medicine, St. Olavs University Hospital, Trondheim, Norway}
\affil[8]{KG Jebsen Center of Exercise in Medicine, Department of Circulation and Medical Imaging, Faculty of Medicine and Health Science, Norwegian University of Science and Technology, Trondheim, Norway.}
\affil[9]{Department of Cardiology and Center for Cardiological Innovation, Oslo University Hospital, Rikshospitalet, Oslo, Norway}

\affil[10]{Department of Otorhinolaryngology, Head \& Neck Surgery, Oslo University Hospital, Rikshospitalet, Oslo, Norway}

\begin{document}

\maketitle

\begin{abstract}
In this work, we present  an  approach for  unsupervised domain adaptation (DA) with the constraint, that the labeled source data are not directly available, and instead only   access to a classifier trained on the source data is provided.  Our solution,  iteratively labels only high confidence sub-regions of the target data distribution, based on the belief of the classifier. Then it  iteratively learns new classifiers from the expanding high-confidence dataset.   The goal is  to apply the proposed approach on DA for the task of sleep apnea detection and   achieve personalization based on  the needs of the patient.  In a series of  experiments with both open and closed  sleep monitoring datasets, the proposed approach is applied  to data from different sensors,  for DA between the different datasets.  The proposed approach outperforms in all experiments  the classifier trained in the source domain, with an improvement of the kappa coefficient  that varies from 0.012 to  0.242. Additionally,  our solution is applied to  digit classification  DA  between  three well established digit datasets, to investigate the generalizability of the approach, and to allow for comparison with  related work. Even without direct access to the source data,  it  achieves good results, and outperforms several well established unsupervised DA methods.
\end{abstract}

\section{Introduction}
In the context of Machine Learning, Domain Adaptation (DA) has recently  been successfully applied to address the problem of shift across different domains.  DA aims to improve learning for a predictive task at a target, assuming  a source and a target domain  for which  the  distributions of source and target  differ   \cite{pan2009survey,ganin2014unsupervised}. The predictive tasks are the same across the two domains. A sub-field of DA is unsupervised DA, for which  labels are provided only for the source data \cite{kang2019contrastive}, but not for the target data.

However, there are use-cases in which the source data cannot be used together with the target data.  Consider  in a medical setting a scenario in which  a patient collects  health data with a smartwatch.  Health data is typically examined by a medical expert to detect a certain health condition or its absence. Recent research results have shown that Machine Learning can successfully be used for such  tasks like sleep apnea detection \cite{kristiansen2018data,nikolaidis2019augmenting,xie2012real}.  However, training to create such a classifier can be difficult  because, in a supervised setting,  experts are required to label the data and the classifier needs to be personalized for each individual. The latter is necessary because individuals are different in terms of physiology, prevalence, sensor placement and  the same signals might be measured with different sensors, for example different smartwatch brands.

In our lab, we face a similar scenario. We aim to develop machine learning solutions for sleep monitoring to identify signs of sleep apnea. Sleep apnea is a severe disorder that is rather common, but unfortunately strongly under-diagnosed. To enable “anybody” to use such solutions at home, the data shall be collected with low-cost consumer electronics, including smart phones and smart watches. We use as foundation for our research data from a large clinical study, called A3 study, at the Oslo University Hospital and St. Olavs University Hospital. In this study, sleep monitoring data from several hundred patients is collected and analyzed. A portable sleep monitor certified for clinical diagnosis (i.e., Nox T3 ) has been used for data acquisition. Currently, we achieve with a CNN trained on this data a classification accuracy of approximately 80\%. However, we cannot guarantee that this model can generalize well to new data from an end-user, because of potential domain shifts. Such domain shifts can for example be caused by characteristics of individual end-user sleep data that are not represented in the A3 dataset and quality issues. The latter is based on the fact that a sleep monitor that is certified for clinical diagnosis produces data with substantially higher quality than that of data collected by end-users at home with consumer electronics. Existing Domain Adaptation solutions cannot be applied to create a classifier that is adapted to the end-user data, because (1) end-users might not want to give us their data for privacy reasons and we do not have the resources to support many end-users, and (2) privacy regulations do not permit to share the A3 dataset (e.g., with end-users or third parties).

This scenario can be generalized. Assume a host which has trained a model with labeled data for classification, but the model cannot be directly used on data collected by an end-user due to the presence of domain shift. Both entities, i.e., host and end-user do not want (or cannot) share their data with the other entity.    Thus, the only way to create a personalized classifier  for the end-user is DA without direct access to the labeled source data.  Furthermore,  it must be considered that the end-user potentially has less computing resources than the host (e.g., lack of dedicated hardware components, lack of sufficient GPU memory  to load the data, etc). As a result even if the host is willing to share her data, performing joint training on the end-user could be problematic.

To address these issues, we introduce \textit{Step-wise Increasing target Dataset Coverage based on high confidence ($SICO$)}   to efficiently perform DA at the end-user.  $SICO$  performs unsupervised  DA with the use of only a classifier trained on  the source domain and unlabeled data from the target domain. One of the fundamental ideas in this work is to release only the classifier trained on the source domain, since extracting personal information from a classifier, especially for time series data, is harder than having direct access to the true data of the individual.\footnote{However it is not impossible to extract private information from a model \cite{Fredrikson:2015:MIA:2810103.2813677}. As on-going work we investigate to introduce privacy guarantees in the form of differential privacy.} At the same time,  since the training will be done at the end-user, we take into account possible lack of hardware resource capabilities. By using only a single trained classifier and only the data from the end-user, the proposed approach  is  less resource intensive  than normal unsupervised DA.

Based on this, the contributions of this work include: (1) the introduction of $SICO$,  a  DA technique which leverages the neuronal excitation of the output neurons as a means to iteratively select high confidence regions to train with. The goal of this process is to incrementally address  the existing domain shift, and  generalize well to the  data of the end-user. (2) The application of the proposed approach on the real-world problem that we are interested in, namely sleep apnea detection and  the investigation of the effectiveness for different physiological sensors and datasets. (3) The investigation of the proposed approach to perform DA for a different type of data and  for a different task (i.e., digit classification), which showcases its generalizability.

\section{Method}
\label{sec:Method}
In this section, we describe the proposed approach, and  provide some insights about  how and why it works. 

\subsection{Step-Wise Increasing Target Dataset Coverage based on High Confidence}

\begin{figure*}[h]
\centering
  \includegraphics[scale=0.3]{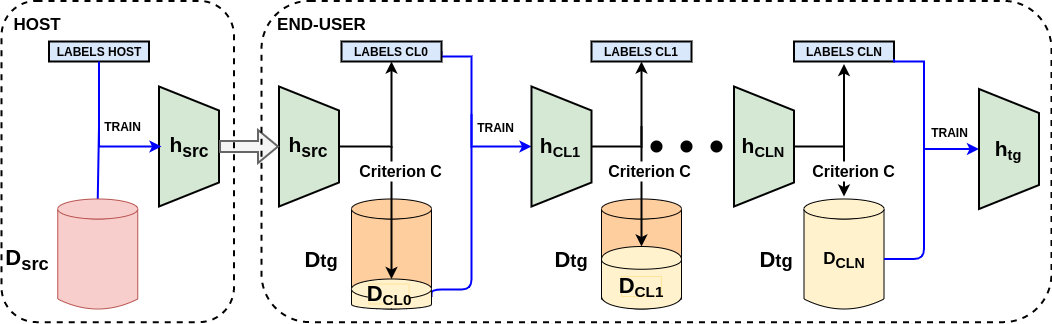}
\caption{ In the proposed method, we train $h_{src}$ on $D_{src}$, and  we release it to the end-user (target). The end-user incrementally adapts  new classifiers to its domain based on the iterative beliefs of $h_{src},h_{CL1}...h_{CLn}$ and a criterion $C$.}
\label{fig:Method}
\end{figure*}

 In this work, we assume that  $h_{src}$ is  trained with data from a host with a labeled dataset $D_{src}$ (see Figure \ref{fig:Method}), and afterwards is released to the public. An end-user   has  an unlabeled dataset $D_{tg}$ and wants to classify $D_{tg}$ on the same \textit{task} \cite{pan2009survey} that $h_{src}$ has been trained for. The goal of $SICO$ is, given $h_{src}$, to create a new classifier  $h_{tg}$ that is adapted to $D_{tg}$.

\par   We assume that  $h_{src}$ is a neural network which performs density estimation of the true conditional distribution $p(y|x)$.  The core idea of the proposed approach (see Figure\ref{fig:Method}) is to start with the data of the host $D_{src}$, and then train and release $h_{src}$ with $D_{src}$. Then,  assuming that the end-user has access to an unlabeled dataset $D_{tg}$,  we select the subset of $D_{tg}$  that  satisfies a criterion $C\{h_{src}\}$,  which we will call $D_{CL0}$. This subset consists of  the data for which, depending on the formulation of $C$, $h_{src}$ is highly confident about their true label. Thus, we use $h_{src}$ to label $D_{CL0}$, and train a new classifier $h_{CL1}$ on $D_{CL0}$ with these labels. We repeat the procedure  with  $h_{CL1}$ labelling  from the remaining data  of $D_{tg}-D_{CL0}$  in order to create a new dataset $D_{CL1}$  consisting of $D_{CL0}$ together with all the data that satisfy $C\{h_{CL1}\}$ from $D_{tg}-D_{CL0}$. We then train $h_{CL2}$ from $D_{CL1}$. We repeat these steps until a terminal condition is being met.   In algorithmic form,  the method includes the following steps:

\begin{itemize}
\item \textbf{Step 1:} We train $h_{src}$ on $D_{src}$ and release $h_{src}$ to the public
\item \textbf{Step 2:} Based on a given confidence criterion $C\{h_{src}\}$ (for example that the logits of the classifier  for a  class are larger than a threshold $T$ ) choose a subset $D_{CL0}\subset D_{tg}$ such that  $C\{h_{src}\}$ is satisfied $\forall \textbf{x}^i_T \in D_{CL0}$.
\item \textbf{Step 3:}  Based on the labels $Y_{CL0}$ that  $h_{src}$ produces for $D_{CL0}$  train a new classifier  $h_{CL1}$ with $D_{CL0}$ and $Y_{CL0}$
\item \textbf{Step 4:} Set $i=1$. Repeat \textbf{Steps 5, 6} and \textbf{7} until   $D_{CLi}=D_{tg}$ ,  or until a terminal condition is  met (e.g., $i\leq N$).
\begin{itemize}
    \item \textbf{Step 5:} Based on $C\{h_{CLi}\}$ choose  $D_{CLi}$ such that $D_{CL(i-1)}\subset  D_{CLi}\subseteq D_{tg}$. $D_{CLi}$ is defined as the subset: $D_{CLi}=$(($D_{tg}-D_{CL(i-1)}$) s.t   $C\{h_{CLi}\}$) $\cup D_{CL(i-1)}$. Thus, $D_{CLi}$ contains all $D_{CL(i-1)}$ together with all the datapoints of $(D_{tg}-D_{CL(i-1)})$ which satisfy $C\{h_{CLi}\}$. 
    \item \textbf{Step 6:} Use $h_{CLi}$ to label $(D_{CLi}-D_{CL(i-1)})$. Unite with $Y_{CL(i-1)}$ to create $Y_{CLi}$ 
    \item \textbf{Step 7:} Based on the labels $Y_{CLi}$  that $h_{CLj},  j=src,1...,i$  have produced for $D_{CLi}$ train a new classifier  $h_{CL(i+1)}$ with $D_{CLi}$ and $Y_{CLi}$. Set $i=i+1$.
\end{itemize}
\item \textbf{Step 8:}  Return $h_{tg}=h_{CLi}$
\end{itemize}
Please note that choosing a good criterion $C$ is vital for the success of the algorithm. If we  choose improper $C$ and we have a demanding terminal condition to meet (like having a high threshold of belief $\forall \textbf{x}_T^i\in D_{tg}$), the algorithm could  "get stuck", or the performance could suffer. In the next subsections we discuss potential choices for $C$.  
\subsection{Methodological Analysis }
We need two core characteristics  for the method to work well: (1)  We need  $h_{src}$ to be sufficiently trustworthy such that if we satisfy $C\{h_{src}\}$ for a given $\textbf{x}_T^i$ (a datapoint from the input space), there is a high probability that $\textbf{x}^i_T$ belongs to the true  class that $h_{src}$ predicts. Thus, if the data distributions of the host and the end-user are very different,  it is clear that we cannot expect very good performance either from $h_{tg}$, or from $h_{src}$. This observation is in-line with the theoretical analysis for domain adaptation from \cite{ben2007analysis} (see Theorem 1). (2) We need  $h_{CLi}$ to be trained on a subset of $D_{tg}$ with labels for  which we are confident to generalize to new data from $D_{tg}$ with high confidence. This is equivalent to a classifier  generalizing well to new data. Thus,  the design of  $h_{CLi}$ needs to be good  enough, and also the dataset $D_{CL(i-1)}$  with which $h_{CLi}$ has been trained should be large enough to give $h_{CLi}$ the ability to generalize well. This characteristic, i.e., (2) is  needed for all the classifiers during the algorithm. 

\par  At the last step of the algorithm, the empirical risk  of  $h_{tg}$  (for cross-entropy loss)  for the last  dataset $D_{CLn}\subseteq D_{tg}$ is:

$$ \hat{L}(h_{tg})=-\frac{1}{|D_{CL_n}|}\sum_{\mathbf{x}_T^j\in D_{CLn}} \sum_c y^c_{j,CLi}log(h^c_{tg}(\textbf{x}_T^j))$$
 where    $y^c_{j,CLi}$  is the element $c$ of the one-hot encoded-vector of $argmax_c\{h^c_{CLi}(\textbf{x}^j_T)\}$, with $i \in \{0,n\}$ ,the index for the classifier which labelled $\mathbf{x}_T^j$, and 0 referring to $h_{src}$. Notice that it is not necessary that $D_{CLn}=D_{tg}$ as we could stop the algorithm before the criterion holds for the whole $D_{tg}$. We have for the true empirical risk of $h_{tg}$, with the true labels:
\begin{equation}
\begin{split}
 &L(h_{tg})=-\frac{1}{|D_{CL_n}|}\sum_{\mathbf{x}_T^j\in D_{CLn}} \sum_c y^c_{j}log(h^c_{tg}(\textbf{x}_T^j))\\
 &= -\frac{1}{|D_{CL_n}|}\sum_{\mathbf{x}_T^j\in D_{CLn}} \sum_c y^c_{j,CLi}log(h^c_{tg}(\textbf{x}_T^j)) - \frac{1}{|D_{CL_n}|}\sum_j \delta_j \cdot log(h_{tg}(\textbf{x}^j_T))\\
 &=\hat{L}(h_{tg})+ \frac{1}{|D_{CL_n}|}\Delta(h_{tg})
\end{split}
\label{eq:basic}
\end{equation}
with   $\delta_j=y_j-y_{j,CLi}$,  and  $y_j$   the true label vector of $\textbf{x}^j_T$. Notice that when $y_j=y_{j,CLi}$, $\delta_j=\mathbf{0}$. Intuitively, $\Delta$ can be thought as a form of accumulative error of the algorithm, and the larger it is the bigger the difference of the true loss and the actual loss we are minimizing. This obviously has a negative impact in the performance on new data.

Additionally, notice that from the above algorithm, in  the labels $Y_{CLi}$ for a dataset $D_{CLi}$ there are as many errors expected as there are made from $h_{CLi}$'s generalization, plus the errors that were "passed" from the generalizations of the previous classifiers. However, $h_{CLi}$'s  generalization error is also dependent  on the generalization error from the previous classifiers, because it uses $D_{CL(i-1)}$ for training with the labels of the previous classifiers. Based on this discussion, and assuming that we use $n$ classifiers, $\Delta$ can be rewritten recursively in the following form:
\begin{equation}
\begin{split}
&\Delta(h_{tg})=\Delta_n(h_{tg};h_{src},h_{CL1}...h_{CLn})\\
&=-\sum_{\mathbf{x}^j_T : y_j\neq y_{j,CLi}} \delta_j \cdot log(h_{tg}(\textbf{x}^j_T))\\
&=-\sum_{\substack{\mathbf{x}^j_T\in (D_{CLn}-D_{CL(n-1)})}}\delta_j(h_{src},h_{CL1}...h_{CLn})\cdot log(h_{tg}(\textbf{x}^j_T))\\
&+ \Delta_{n-1}(h_{tg};h_{src},h_{CL1}...h_{CL(n-1)})\\
\end{split}
\end{equation}
\label{eq:delta}
where we define as $\Delta_0(h_{tg};h_{src})= -\sum_{\mathbf{x}^j_T\in D_{CL0}}\delta_j(h_{src})\cdot log(h_{tg}(\textbf{x}^j_T))$. From this form it is straightforward that the earlier classifiers play a more important role in the performance of $h_{tg}$, since their error in a sense "propagates" through the next training iterations. Thus, $h_{src}$ plays the most important role as the first classifier in the algorithm. 

Furthermore,  for every step $i$ of the algorithm,   the  generalization capability of the classifier  $h_{CLi}$ plays an important role in the accumulative $\Delta$.  If $h_{CLi}$'s generalization capability  is low, it will assign many erroneous labels during the labeling of its high confidence dataset for criterion $C\{h_{CLi}\}$. This will also affect the next steps, since we have  to expect worse generalization capability for the next classifiers, because these data will become training data to the next classifier etc. Finally, notice  that the above analysis  can be applied to any $h_{CLi}$ and its respective   $D_{CL(i-1)}$ instead of $h_{tg}$ and $D_{CLn}$.

\subsection{Choosing the Belief  Criterion }

An essential part of the proposed method is the criterion $C$ with which we choose for a classifier $h_{CLi}$ the new subset of $D_{tg}$ to include to  $D_{CL(i-1)}$, i.e., $(D_{CLi}-D_{CL(i-1)})$.   We  label this part with $h_{CLi}$, and then combine it with $D_{CL(i-1)}$  and use to train $h_{CL(i+1)}$. In our case,  we assume that all $h_{CLi},h_{src},h_{tg}$  are neural networks with softmax output.  We therefore take advantage of the neuronal excitation  of the output class neurons. We use neuronal excitation as an indication of confidence  that $\textbf{x}^i_T$ belongs to a class. 
 
 There are many ways we can use the excitation of the output neurons as a criterion to assign labels for a high confidence sub-dataset. Examples include thresholding, selecting the $m$ datapoints which  lead to the strongest activation for each class neuron, or  if the class of an output neuron becomes the maximum class for $m$ datapoints, pick the $m/100$ datapoints that activate this neuron the most, etc.  

\subsection{$SICO$ and Curriculum Learning}

At first glance $SICO$ and Curriculum Learning (CL)  \cite{bengio2009curriculum} can appear to be quite similar. In CL the training "starts small" by using a small training set with easy examples identified with the use of a scoring function \cite{hacohen2019power}. Afterwards, CL  progressively utilizes more difficult examples which are added to the curriculum. Similarly, we  utilize easier examples in terms of domain similarity, and progressively train with  harder datapoints. The previous classifier's class probabilities give us a measurement of the datapoints that are easier for the classifier in terms of the classifiers' higher confidence regarding these points (lower entropy). We hypothesize that datapoints that are easier -in terms of lower entropy- for a classifier trained in a different domain are more likely to be  more similar to datapoints from the source domain, in terms of class separation.

However, the basic vanilla CL uses a static scoring function  during training. $SICO$ utilizes instead a sequence of  scoring functions (i.e., $h_{CLi}$) that are learned dynamically as the training process continues. The creation of a new  scoring function  depends on the dataset and the previous scoring function. Additionally, the first scoring function, i.e., $h_{src}$,  is independently trained and acts as a Teacher that provides the initial scoring paradigm, in conjunction with the belief criterion used. We compare our method with other more relevant and recent works of CL for DA in Section \ref{sec:RelatedWork}.  

\section{Experiment Description}
\label{sec:ExpDescr}
The main goal of this work is to perform successful domain adaptation for bio-sensory signals for the purpose of sleep apnea detection. Additionally, we complement the physiological datasets with datasets for digit classification, i.e., USPS, MNIST, SVHN for two reasons. First, the majority of the related literature uses these datasets  and  so they serve as a baseline for comparison. Second, we want to investigate the generalizability of our approach for different types of data. 

In the next subsections, we describe the datasets used and discuss the experimental set-up.

\subsection{Datasets}
We use the following six datasets to evaluate $SICO$:

\begin{itemize}
\item \textbf{MNIST} \cite{lecun1998gradient}  ($M$) is a dataset containing 60000 28$\times$28  images of digits (handwritten black and white images of 0-9) as a training set.  The test set comprises of 10000 images.
\item  \textbf{USPS} \cite{hull1994database}  ($U$) is another handwritten digit   dataset  (0-9),  which contains 7291 grayscale 16$\times$16 training and 2007 test images.   
\item  \textbf{SVHN} \cite{netzer2011reading}:  ($S$) is s a real-world image dataset  obtained from house numbers in Google Street View images. Similarly to the previous datasets, classification is performed for digits 0-9. It contains 73257 digits (32$\times $ 32 colored images) for training, 26032 digits for testing, and 531131 additional training data. We use only the original training dataset of 73257 digits.
\item  \textbf{Apnea-ECG}  \cite{penzel2000apnea}  ($AE$) is an open dataset from Physionet, containing  sensor data from chest respiration, abdomen respiration, nasal airflow (NAF),   oxygen saturation and Electrocardiograph (ECG). $AE$ has been used in the Computers in Cardiology challenge \cite{penzel2000apnea} and it contains high quality data. It has been collected with Polysomnography in a sleep laboratory. From the 35 ECG recordings in the dataset, 8 recordings (from 8 different patients) contain data from all the sensors. Each recording has duration of 7-10 hours. The sampling frequency  of all sensors is 100Hz, and labels are given for every one-minute window of breathing. The labels identify which minutes are apneic and which are not (i.e., if a person experiences an apneic event during this minute). From $AE$, we use  the NAF, chest respiration, and oxygen saturation signals.
\item  \textbf{MIT-BIH}  \cite{ichimaru1999development}  ($MB$) is an open dataset containing  recordings from 18 patients. The recordings contain different  respiratory sensor signals.  In 15 recordings, the respiratory signal  has been collected  with NAF.  For this reason, we focus on the NAF signal for the $MB$ dataset.  Due to   misalignment of the different signals and lack of labels for the apneic class in 4 recordings, we utilize 11 of the 15 recordings (slp60,slp41 and slp45 and slp67x are  excluded). It is important to note that the data/labelling quality of the $MB$ dataset is  low, which leads to low  classification performance for $MB$ compared to the other respiratory datasets that we investigate \cite{kristiansen2018data}. The labels are given for every 30 seconds and the sampling frequency of all sensors is 250Hz.
\item 
The \textbf{A3 study} \cite{traaen2019treatment} ($A3$) investigates the prevalence, characteristics, risk  factors and type of sleep apnea in patients with paroxysmal atrial  fibrillation. The data were obtained with the use of the Nox T3 sleep monitor with mobile sleep monitoring at home, which in turn  results into lower data quality than data from polysomnography in  sleep-laboratories. An experienced sleep specialist scored the recordings manually using Noxturnal software such that the beginning  and end of all types of apnea events is marked in the time-series  data. To use the data for the experiments in this paper, we labeled  every 60 second window of the data as apneic (if an apneic event  happened during this time window) or as non-apneic. The data we use in  the experiments is from 438 patients and comprises 241350 minutes of  sleep monitoring data. The ratio of apneic to non-apneic windows is  0.238. We use only the NAF signal from the A3 data in the experiments,  i.e., the same signal we use from Apnea-ECG.

\end{itemize}

\subsection{Preprocessing}

The data in  all  sleep apnea datasets is standardized (per physiological signal), downsampled to 1Hz and the windows are shuffled randomly. The data from the $A3$ study,  are very unbalanced, i.e., it contains many more  non-apneic than apneic one-minute windows. Therefore, we rebalance the dataset to contain equal amount of apneic and non-apneic one-minute windows. Since the labels in $MB$  are given every 30 seconds, while $AE$ and $A3$ are labeled every 60 seconds,  we adapt the labelling in $MB$ to 60 seconds by using the following rule: if both 30 second labels are non-apneic then the 60 seconds label is  non-apneic, elsewise it is apneic. For  $A3$, we use 80\% of data as training and 20\% as test set.    For $AE$ we use 25\% of the data as test set, and for $MB$ we use  15\% of the data as test set. We use less   test set  data for $MB$ because  we want to utilize more data for training due to the low quality.

We rescale the data in all digit classification datasets from 0-255 to 0-1. Additionally, for $U$ and $S$, we restructure the data so that  it is in similar form to the $M$ data. We up-scale  the images in $U$ from 16$\times$ 16 to 28$\times$28, and we downscale the images in $S$ from 32$\times$32 to 28$\times$28. Additionally,  we convert the color images in $S$ to grayscale images.

\subsection{Experimental Set-Up}

  We follow in the experiments the steps outlined in Section \ref{sec:Method}, i.e., we train $h_{src}$ such that it can generalize well for the test set of $D_{src}$ and  release $h_{src}$ to the end-user. Then we use $h_{src}$ and a subset of $D_{tg}$ to iteratively create $h_{tg}$. The performance of $h_{tg}$ is evaluated with the test set of $D_{tg}$. This means that we evaluate the proposed method on the test set of $D_{tg}$. We  use the convention $D_{src}\rightarrow D_{tg}$ to indicate that $h_{src}$ is trained on $D_{src}$ and $h_{tg}$  is evaluated on $D_{tg}$.  For each experiment we describe the core algorithmic decisions (e.g. the belief criteria). Further details about other minor algorithmic decisions can be found in Appendix A. 
 Since $D_{tg}$ is not labeled,  we do not have access to a validation set during the training of the proposed method. Thus,  we train each classifier $h_{CLi}$ for a fixed number of batch iterations.

 \subsubsection{Belief Criteria} We empirically investigated to use a fixed threshold for the choice of high confidence data. However, this did not perform well, since the trained classifiers which are transferred from the source domain have biases towards certain classes in the target domain. This results in unbalanced  $D_{CLi}$, which negatively affects the performance.  For this reason, we select the $m$ datapoints per-class which excite the class output neurons the most as belief criterion $C\{h\}$ in all experiments (except $M\rightarrow U$).  We use this criterion in order to maintain the class balance since all datasets with the exception of $U$ that are used as $D_{tg}$ are relatively well-balanced. Since $U$ is not well balanced, we choose a criterion that gives more "freedom" to the classifier to perform the balancing.  Instead of choosing an absolute number N, as $C\{h\}$, we select a percentage $p$ of the datapoints which activate a class output neuron the most.

  \section{Experiments and Results}
\label{section:ExpEval}

The first set of experiments evaluates $SICO$ for digit classification and investigates  three combinations that are commonly used in literature, i.e., $M\rightarrow U$, $U\rightarrow M$, $S\rightarrow M$. The second set of experiments focuses on  sleep apnea detection with physiological sensors.   The  application scenario that we are interested in focuses  on  the transition from high quality sensor data to low quality sensor data. Additionally, we investigate combinations for which the low quality data are used as $D_{src}$, in order to get a more complete picture of the capabilities of the proposed  algorithm. We focus on the NAF signal for the combinations $A3\rightarrow AE  ,A3\rightarrow MB$ ,  $MB\rightarrow AE$, $AE\rightarrow MB$, $AE\rightarrow A3$, and evaluate the abdominal respiration, the chest respiration and the oxygen saturation sensor signals for the $A3\rightarrow AE$ and  $AE\rightarrow A3$  combinations.

\begin{figure}[h!]
\includegraphics[width=12.5cm,height=7.5cm]{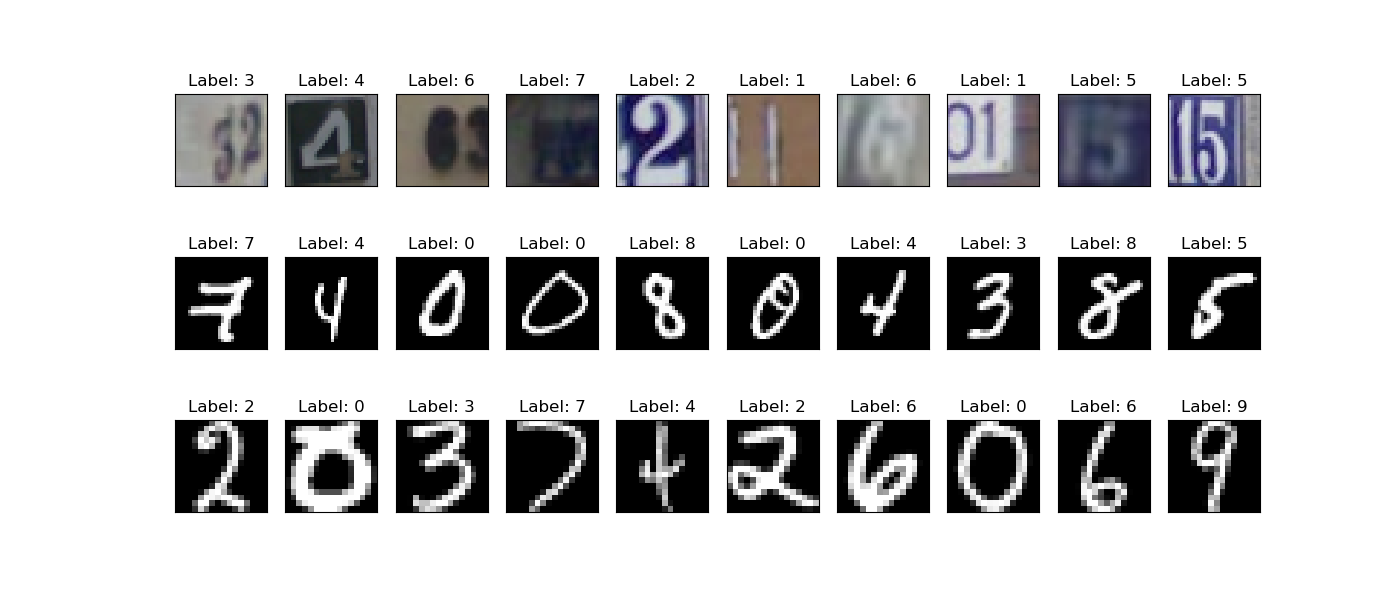}
\caption{Examples of different datapoints from the Digit datasets: First row: examples from SVHN. Second row: examples from MNIST. Third row: examples from USPS.  }
\label{fig:SMU}
\end{figure}

\begin{figure}[t]
\includegraphics[width=12.5cm,height=8.5cm]{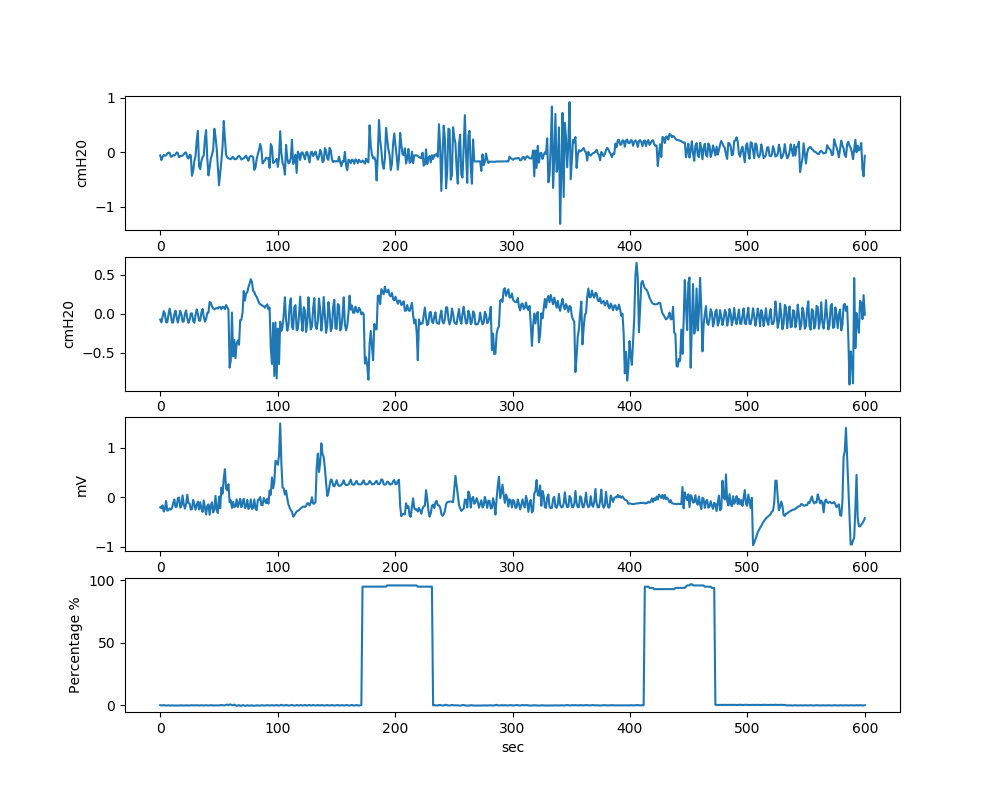}
\caption{Different respiratory signals from $MB$ and $AE$. All rows show randomly chosen 600sec windows of respiratory signals from $MB$ (first row with Resp N) and $AE$ (Second Row with Resp N. Third row with Resp C. and Fourth row with Sp02.   }
\label{fig:MBandAE}
\end{figure}

Figures  \ref{fig:SMU} and \ref{fig:MBandAE} show examples from the different datasets (before pre-processing). From the $MB$ dataset we use only the NAF signal in our experiments. For this reason, we include only the NAF signal from $MB$ in Figure \ref{fig:MBandAE}. It is difficult to visually assess the respiratory signals and extract the domain specific features, especially since the variance of the data is very high. This is apparent in Figure \ref{fig:MBandAE} when trying to compare between the NAF data from $AE$ and $MB$. Generally, though data from $MB$ seem to have higher variance than the data from $AE$ between apneic and non-apneic periods, and also higher fluctuations in the breathing pattern of the patients. Regarding the digit datasets, the differences between the datasets are more apparent. For example some obvious differences we can identify from Figure  \ref{fig:SMU} are: (1) $M$ and $U$ are handwritten digits, whereas $S$ are artificially made, (2)  $S$ in many cases contains additional numbers in the image, (3) digits from $U$ seem to capture larger part of the image in relation to digits from $M$.

\subsubsection{Devices} Finally, please  note that for the sleep apnea set of experiments,  the devices used are of the same type across the different datasets ($AE,MB$ and $A3$) i.e., nasal thermistors for the NAF signal, Respiratory Inductive Plethysmography (RIP) (Chest-Abdomen belts) for the Resp A and C signals, and pulse oximeter for the  measurement of oxygen saturation Sp02.

\subsection{Metrics}
For the  Digit Classification experiments, we use accuracy as   performance metric  since it is commonly used in related literature for the particular task, assuming a well-balanced dataset. For the  apnea detection experiments, we use the kappa  coefficient \cite{cohen1960coefficient} as performance metric since it  better  captures  performance characteristics in a single metric than accuracy, as it takes into account the possibility that two annotators (real labels and predictions in our case) agree by chance.  For completion, we present the accuracy, specificity, and sensitivity results in Appendix B. All experiments are  repeated for  5 iterations, and we present the average results  and the standard error.

\begin{table}[h!]
\centering
\caption{ Architectures used for the experiments (Conv: input channels$\times$output channels$\times$filter, MP: Max Pooling, fc: Fully connected, input$\times$output)}

\begin{tabular}{ | p{2.5cm}| p{2.5cm}| p{2.8cm}|p{2.5cm}|}
\hline
\multicolumn{4}{|c|}{Architectures used} \\
\hline
Layers &$M\leftrightarrow U$&$S \rightarrow M$ &Sl.Apnea\\
\hline
Conv+MP&$1 \times 32\times 5\times5$&$1 \times 32\times 5\times5$& $1\times 16 \times 4$ \\
Conv+MP&$32\times 28\times5\times5$&$32\times 64\times 3\times 3$&$ 16\times 32\times 4 $\\
Conv(+MP)& $28\times 28\times5\times5$  &$64\times 128\times5\times5$ &$ 32\times64\times 4$ \\
FC&$(7\times 7 \times 28)\times 1024$ &$(7\times 7\times128)\times 3072$ &$ (8\times64)\times 64$\\
FC&$1024\times 128$ &$3072\times 1024$&$ 64\times 32$ \\
out&  $128\times 10$&$1024\times 10$& $32\times2$   \\
\hline
\end{tabular}
 \label{table:Archs}
\end{table}

\subsection{Digit Classification}

We use the same architecture for all classifiers  (i.e., $h_{src}, h_{CL1}...h_{CLn},h_{tg}$) in all Digit classification experiments. This means that for any given instance of the algorithm we have only one model in memory. We use a convolutional neural network (CNN) with a similar  but wider architecture to LeNet-5  with more weights per layer (see Table \ref{table:Archs}), and one more fully-connected layer and Convolution layers. We chose this architecture  as this is a well-established simple model that is very often  used for  digit classification.  Note that  we do not use MaxPool. on the third Conv. layer for the  $S\rightarrow M$, and $M\leftrightarrow U$ experiments. We use more weights to potentially compensate for the larger datasets (i.e., SVHN) and  relu activations to all layers, softmax output, and dropout in the   fully-connected layers. Additionally for the network of the $S\rightarrow M$ experiment we perform  Batch Normalization. For all experiments,  we use  a batch size of 128, learning rate of 0.001, and the Adam optimizer \cite{kingma2014adam}.  All differences in results (mean accuracies) between $h_{src}$ on the $D_{tg}$ test set and  $h_{tg}$ on the  $D_{tg}$ test set are statistically significant based on the one-tailed paired t-test (for $p=0.05$). We use this test as an indication of the importance of the improvement  in performance that we observe for $h_{tg}$ compared to $h_{src}$ on the $D_{tg}$ test set.\\

 We trained $h_{src}$  for $M\rightarrow U$ and $U\rightarrow M$ for  4688 batches (i.e., 600K datapoints). For $S\rightarrow M$, we trained $h_{src}$ for  7812 batches (i.e., $10^6$ datapoints), since $h_{src}$ does not converge with only 4688 batches when trained on $S$.\\
When $D_{tg}$ is either $M$ or $U$ we use  the fixed number of datapoints criterion $C\{h\}$ as explained in Section \ref{sec:ExpDescr}.3.1, with $m=200$ datapoints per-class to construct $D_{CL0}$, and $m=100$ datapoints per-class for all subsequent $D_{CLi}$. The algorithm stops when we do not have any more unlabeled data in $D_{tg}$. For more details please refer to Appendix A.\\

\begin{table}[h!]
\caption{Accuracy of digit classification }
\centering
\begin{tabular}{ | p{1.70cm}| p{1.8cm}|p{1.8cm}|p{1.8cm}|}
\hline
\multicolumn{4}{|c|}{ $SICO$ $h_{tg}$  performance on digit classification DA (Accuracy)} \\
\hline
 & $M\rightarrow U$ & $U\rightarrow M$  &$S\rightarrow M$ \\
\hline
$h_{src}$,$D_{src}$:&99.12$\pm$0.01&96.62$\pm$0.16&90.55$\pm$0.40\\
$h_{src}$, $D_{tg}$: &79.83$\pm$0.51&69.58$\pm$2.00&65.94$\pm$1.97\\
$h_{tg},D_{tg}$:&89.32$\pm$0.70&90.88$\pm$0.69&86.95$\pm$1.77\\
\hline
\end{tabular}
 \label{table:Digits}
\end{table}

Table \ref{table:Digits} presents the classification performance of the three combinations for $h_{src}$ on $D_{src}$, and $h_{tg}$ and $h_{src}$ on $D_{tg}$.  From Table \ref{table:Digits}  we notice that $h_{tg}$ outperforms $h_{src}$ for all $D_{src}\rightarrow D_{tg}$ cases. This could potentially be attributed to the use of domain specific knowledge (in the form of training data from $D_{src}$)  to train  all subsequent $h_{CLi}$ and the $h_{tg}$ , with a given confidence defined by the used criterion. Notice the steep drop of $h_{src}$ in all cases from $D_{src}$ to $D_{tg}$,  which are expected due to the domain differences. We notice the largest drop for the case of $S\rightarrow M (24.61\%)$. The largest improvement is observed again  for the case of $S\rightarrow M$ (21.01\%).  \\

\subsection{Sleep Apnea Detection}

To perform sleep apnea DA,  we use again identical architectures for all classifiers, i.e., $h_{src}, h_{CL1}...h_{CLn},h_{tg}$. We use a 1D CNN (see Table \ref{table:Archs}), and use relu activations, dropout on the fully connected layers, and softmax activations on the output. When the $A3$ study is $D_{src}$, we train $h_{src}$ for 15 epochs and when $MB$, or $AE$ are $D_{src}$ we train $h_{src}$ for 20 epochs (in order to have more training iterations since $MB$ and $AE$ are  smaller than $A3$). We use  in all experiments a batch size of 128, learning rate of  0.001, and  the Adam optimizer \cite{kingma2014adam}. Note that  all differences in results  between $h_{src}$ on the $D_{tg}$ test set and  $h_{tg}$ on the  $D_{tg}$ test set are statistically significant based on the one-tailed paired t-test (for $p=0.05$), with  the exception of Resp A: $A3 \rightarrow AE$ ,  NAF: $A3 \rightarrow MB$, and Resp C: $AE\rightarrow A3$. \\

We use in all experiments with $AE$ and $MB$ as $D_{tg}$    the fixed data criterion with 500 datapoints per-class for $h_{src}$ and 200 datapoints per class for all subsequent $h_{CLi}$  as belief criterion. With $A3$ as $D_{tg}$, we use the fixed data criterion with 10000 datapoints per-class for $h_{src}$ and 500 datapoints per class for all subsequent $h_{CLi}$, since $A3$ is much larger.  The algorithm stops when we do not have any more unlabeled data in $D_{tg}$.\\

\begin{table}[h]
\centering
\caption{Performance for DA between different dataset combinations for the NAF sensor signal.  }
\begin{tabular}{ | p{2.5cm}| p{1.6cm}|p{1.65cm}|p{1.8cm}|}
\hline
\multicolumn{4}{|c|}{ $SICO$ $h_{tg}$  performance  (kappa $\times$100) for NAF} \\
\hline
NAF: &$h_{src}$,$D_{src}$&$h_{src}$,$D_{tg}$ &$h_{tg},D_{tg}$ \\
\hline
$A3\rightarrow AE$:&69.33$\pm$0.21&67.46$\pm$5.38&84.07$\pm$4.76\\
$A3\rightarrow MB$: &69.33$\pm$0.21&10.26$\pm$1.13&19.30$\pm$1.78\\
 $AE\rightarrow MB$:&94.39$\pm$0.49&11.96$\pm$1.15&13.14$\pm$0.63\\
 $MB\rightarrow AE$:&41.69$\pm$1.60&65.27$\pm$3.03&78.88$\pm$2.25\\
 $AE\rightarrow A3$:&94.39$\pm$0.49&29.67$\pm$1.09&36.68$\pm$2.60\\
\hline
\end{tabular}
 \label{table:ApneaDA}
\end{table}

Table \ref{table:ApneaDA} presents the classification performance of the three combinations for $h_{src}$ on $D_{src}$ and $h_{tg}$ and $h_{src}$ on $D_{tg}$. From Table  \ref{table:ApneaDA} we initially notice the significant impact of the quality of the  datasets on the performance of $h_{src}$. For $MB\rightarrow AE$, the quality of the $MB$  data is low enough that the evaluation of $h_{src}$ on the test set of  $MB$ performs worse than the evaluation of $h_{src}$ in $AE$. Since $AE$ is a high quality dataset, it is easier to have much better performance. This is reflected by  the very big difference of kappa ($\times 100$) between the two datasets for $h_{src}$(i.e., 41.69 vs. 94.39).  For  $A3$, we expected that it would have better transferability to the other datasets since it is much larger, thus covering a wider variety of cases (both from a patient and from a data quality perspective).  Though this holds for the $AE$ case, i.e., $A3 \rightarrow AE$, it is not the case for $MB$, for which ($A3 \rightarrow MB$) $h_{src}$   performs very poorly. It is  noteworthy however that we get significantly better results  with $SICO$ for $A3 \rightarrow MB$ than for $AE\rightarrow MB$. In summary, we identify $D_{src}$'s  data quality  and  variation as two important factors which affect the performance  of $h_{src}$ and $h_{tg}$ for $D_{tg}$\\

Generally, $h_{tg}$ performs for all cases  better than $h_{src}$ on $D_{tg}$. As expected from Eq. \ref{eq:delta}, $h_{src}$ plays a very critical role in the $SICO$ process, and we cannot get very good results if the performance of $h_{src}$ is initially very low. We discuss this characteristic in more detail in Section \ref{sec: Discussion}. Finally, another noteworthy observation is the very large standard error for all cases for $h_{src}$ on $D_{tg}$, and to a lesser extent for $SICO$. The results for $h_{src}$ were not stable among the different iterations of the experiment (for example for $A3 \rightarrow AE$ the range of kappa$\times 100$ values was 54.3-81.9). However $h_{tg}$, consistently outperformed $h_{src}$, and it additionally provided a stabilizing effect, as the results for $h_{tg}$ did not vary as much (with the exception of $AE \rightarrow A3$). \\
\subsubsection{Other Sensors}

We repeat the experiment for the $A3 \rightarrow AE$ combination for the additional respiratory sensors which are included in $A3$ and $AE$, i.e., Abdominal Respiration (Resp A), Oxygen Saturation (Sp02)  and Chest Respiration (Resp C). We do not use $MB$ for these experiments due to the small number of recordings per-sensor and the already very low performance it yields for all experiments even with the  NAF signal, for which we have much more data in comparison to the other signals. The other  parameters are the same as in the previous experiments. \\

\begin{table}[h]
\centering
\caption{Performance of $A3 \rightarrow AE$ and $AE\rightarrow A3$ for  Resp A, Sp02, and   Resp C}
\begin{tabular}{ | p{3.5cm}|p{2cm}|p{2cm}|p{2cm}|}
\hline
\multicolumn{4}{|c|}{  $SICO$ $h_{tg}$ Performance  (kappa$\times$100) for different sensors} \\
\hline
 &Resp A &Resp C&Sp02\\
\hline
$A3 \rightarrow AE$: $h_{src}$,$D_{src}$:&66.74$\pm$0.38&66.91$\pm$0.30&71.84$\pm$0.67\\
$A3 \rightarrow AE$: $h_{src}$,$D_{tg}$: &78.95$\pm$1.92&57.23$\pm$8.38&61.23$\pm$4.44\\
$A3 \rightarrow AE$: $h_{tg},D_{tg}$:&80.68$\pm$0.84&81.47$\pm$1.12&74.23$\pm$2.07\\
\hline
$AE \rightarrow A3$: $h_{src}$,$D_{src}$:&92.31$\pm$0.50&90.97$\pm$0.45&88.93$\pm$0.36\\
$AE \rightarrow A3$: $h_{src}$,$D_{tg}$: &27.35$\pm$1.04&23.07$\pm$1.89&-0.32$\pm$0.00\\
$AE \rightarrow A3$: $h_{tg},D_{tg}$:&48.47$\pm$1.20&26.00$\pm$0.53&19.37$\pm$1.58\\
\hline
\end{tabular}

 \label{table:ApneaOther}
\end{table}

The results are found in  Table \ref{table:ApneaOther}.  We notice that  $h_{tg}$ significantly outperforms $h_{src}$  on $D_{tg}$ (with the exception of Resp A).  As before,  we have with  $A3 \rightarrow AE$   a  very large standard error  (big variation) for $h_{src}$, and $SICO$ seems to have a stabilizing effect on the variation. For these experiments, this phenomenon is more pronounced than for the Resp N. experiment. Again, we observe for Resp A: $A3\rightarrow AE$  the same interesting pattern that we observed for NAF: $MB \rightarrow AE$ , i.e., that $h_{src}$ has a higher performance on the  target test set than the source test set. We hypothesize that this happens for similar reasons as before,  i.e.,  due to the better quality and potential homogeneity of $AE$  relative to $A3$. This hypothesis is strengthened by the fact that for the $A3\rightarrow AE$ adaptation all sensor signals seem to perform relatively well (with the exception of Resp C).  Another interesting characteristic is that Sp02 which has the best performance for $A3$, seems to adapt much worse to the new domain,i.e., $AE$, than Resp A.\\

In the $AE\rightarrow A3$  experiments  $h_{tg}$ outperforms $h_{src}$ again for all cases. However,  this time the variation for $h_{src}$ in $D_{src}$ is smaller than for $h_{tg}$. This could potentially again be attributed to the larger data variety in the A3, which can make   DA more stable regardless of the initialization, and  training randomness for $h_{src}$. Interestingly, $h_{src}$ has seemingly not generalized in the Resp SpO2 experiments. Intuitively, this leads us to assume that $h_{tg}$ should also fail. However, the results show that  to some extent that $h_{tg}$ is still able to learn. For a $h_{src}$ trained with $AE$ on Sp02 and tested on $A3$ Sp02, we observe that for the train set, the  specificity between the pseudo-labels and the true labels is 0.952 and sensitivity is 0.08. For the same $h_{src}$, for $D_{CL0}$, specificity is 0.67,  sensitivity 0.33. We notice that we achieve already with the first $h_{CL1}$ a kappa ($\times 100$) value of 16. We hypothesize, that the performance improvement occurs due to the recognition from $h_{CL1}$ of unique characteristics of each class. This could be  a result of the better balancing between the correct data from each class, together with the potentially uninformative nature of the wrong class data. Due to this uninformative nature,  $h_{CL1}$ does not generalize as much towards a wrong pattern.   Intuitively, this means that $h_{src}$ learned a useful structure of the feature space for both classes. Due to the better balancing of $D_{CL0}$, the "hidden" structure learned for the minority class is uncovered for the classifier trained on $D_{CL0}$, i.e., $h_{CL1}$, allowing it to generalize better to new data.

\subsection{Training Analysis}

In $SICO$, we train sequentially $n$ classifiers, each one to a  high-confidence sub-region of the training dataset, which is a superset of the region that the previous classifier has been trained. Thus, assuming that the accumulative error $\Delta_i(h_{tg};h_{src},h_{CL1}...h_{CL(i-1)})$ for every classifier $h_{CLi}$ in the sequence does not get too large, we expect that   the cross-entropy loss  with the true training labels will become smaller,  because each classifier learns with  an expanding amount of datapoints from the training dataset. \\

\begin{figure}
\raggedright
\includegraphics[width=.49\linewidth,height=4cm]{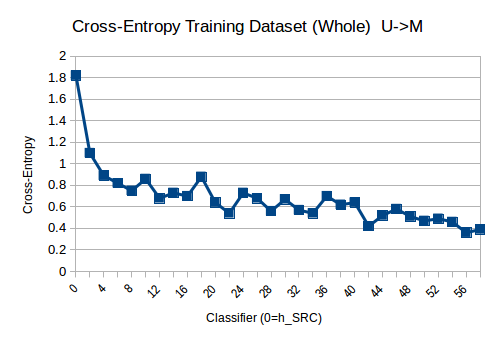}
\includegraphics[width=.49\linewidth,height=4cm]{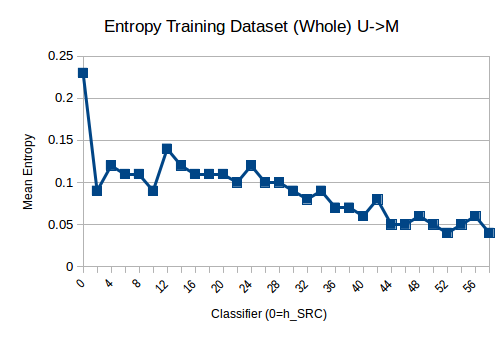}
\label{fig:testa}
\caption{Mean cross-entropy loss with the real labels, (a) and Mean entropy (b) for the whole  target training dataset.}
\label{fig:Graphs1}
\end{figure}

\begin{figure}
\raggedright
\includegraphics[width=.49\linewidth,height=4cm]{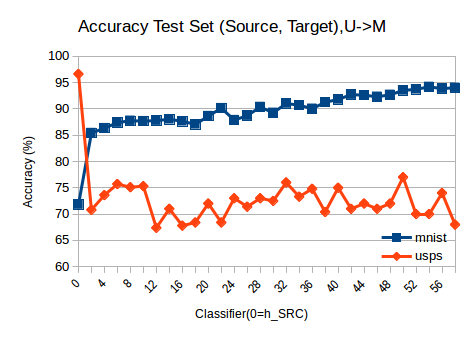}
\includegraphics[width=.49\linewidth,height=4cm]{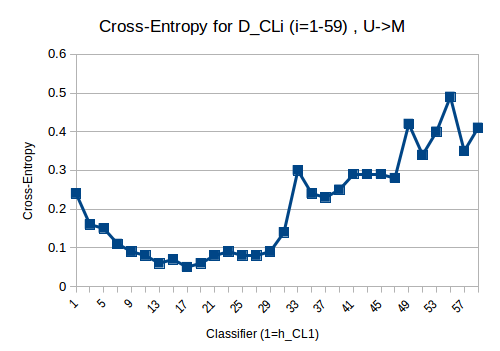}
\label{fig:testa}
\caption{Test accuracy  (hypothetical) for all $h_{CLi}$, (a). Mean cross-entropy loss for each dataset $D_{CL(i-1)}$ for the respective $h_{CLi}$, (b). We show the results for every second classifier $h_{CLi}$ (dots) for  better readability. All figures were obtained on the same run.}
\label{fig:Graphs2}
\end{figure}

 Figure \ref{fig:Graphs1}  (a) presents the training dataset cross-entropy loss for  $U\rightarrow M$, for $m=500$ per-class    for $h_{src}$ and $m=200$ per-class  for the subsequent $h_{CLi}$,  based on the fixed number of datapoints criterion. As expected, the loss decreases  as the datasets $D_{CLi}$ become larger and the algorithm proceeds.  This is mapped also in the performance on new data as  shown in Figure \ref{fig:Graphs2} (a), which depicts the test set performance in terms of accuracy of $M$ for all  different $h_{CLi}$. In this Figure, we also include the performance of  the test set of $U$ (orange) for completion. The performance of $U$ is as expected  degrading as the algorithm proceeds.

Additionally,   Figure \ref{fig:Graphs2} (b)  shows the mean cross-entropy with the real  labels for $D_{CLi}$  for all i. Notice that the error initially becomes smaller, which means that  the new regions that are included in $D_{CLi}$ to form $D_{CL(i+1)}$ yield better performance than the previous i. However, after some iterations this does not hold, and the performance with the new $D_{CL(i+1)}$ degrades in relation to $D_{CLi}$.  This  means that the mean   $\Delta_i$ becomes larger. We hypothesize that this behavior could be attributed to the misalignment between the left out data that later iterations represent and the initial high confidence region of $h_{src}$.\\

 We expect that for  increasing values of i  the average entropy of $h_{CLi}$  becomes smaller. Since each new $h_{CLi}$ uses a larger part of $D_{tg}$ and trains on hard labels from  $h_{CL(i-1)}$,  it will be more "confident" for a larger part of $D_{tg}$.   Figure \ref{fig:Graphs1} (b)  showcases this phenomenon. Furthermore, $h_{CLi}$ has been trained with more data and thus can potentially generalize better to new data.

\section{Discussion}
\label{sec: Discussion}
 From the  above evaluation, we observe that the success of $SICO$ on $D_{tg}$ depends on how well the initial $h_{src}$ can generalize on $D_{tg}$. This observation occurs directly from Eq. \ref{eq:delta}, since the error of $h_{src}$, recursively propagates for all $\Delta_i$ which constitute the total $\Delta_n$.  Thus,   if  $h_{src}$ does not perform well on $D_{tg}$ this has a direct repeated negative impact on the whole $SICO$ algorithm.  Interestingly however, even for the cases for which $h_{src}$ performs really bad on $D_{tg}$ (mainly for the cases which include $MB$ and $A3$ as $D_{tg}$) $h_{tg}$ is still able to outperform $h_{src}$. This  potentially happens due to the fact that we expect better performance for high-confidence regions than the average performance,  assuming that we trust  $h_{src}$ adequately. This means that a smaller error propagates through the algorithm, than for the case in which  we have a lower confidence region. This insight is one of the main inspirations for this work. 
\par However,  if we use too few data points as our confidence region,  then the classifier trained on this region will not be able to generalize well enough to new data, and will   have higher generalization error for the new regions. Thus, there exist a trade-off between confidence and trust on the one hand, and generalization capability on the other hand.  This needs to be taken into account to determine  how strict the criterion $C\{h\}$ should be, as this decides how many datapoints a new $D_{CLi}$ will have.\\
Finally, a natural extension to the proposed approach is to apply probabilistically the criterion $C$ that we use, instead of using it as a hard threshold of acceptance for a given datapoint. As this is a more natural approach that better captures  the probabilistic nature of density estimation that the classifiers perform, we hypothesize that it can yield even better results than the original hard acceptance/rejection method. However, in preliminary experiments  we were not able to showcase improvement in performance in relation to the original method.

\subsubsection{Implications for  Detecting a Condition at the End-User} The proposed approach provides a way for a trained classifier to adapt to a new domain. In the context of healthcare, one application   could regard  adaptation between different patients.  For example, for  the OSA case, if we have a person whose datapoints (i.e., one-minute  windows) are different due to a variety of potential factors than the datapoints $h_{src}$ has been trained with,  we can expect that by applying $SICO$ we can get a better performing classifier for the particular individual. Additionally, assuming that we apply the  criterion $C$, we can properly adapt in a dataset which is balanced with different class frequencies in relation to the source dataset. This, in the context of condition detection means that we could potentially adapt from patients who do not have many pathological datapoints to patients for which the condition is more expressed.

\section{Related Work}
\label{sec:RelatedWork}

\begin{figure}
\centering
 \includegraphics[scale=0.25]{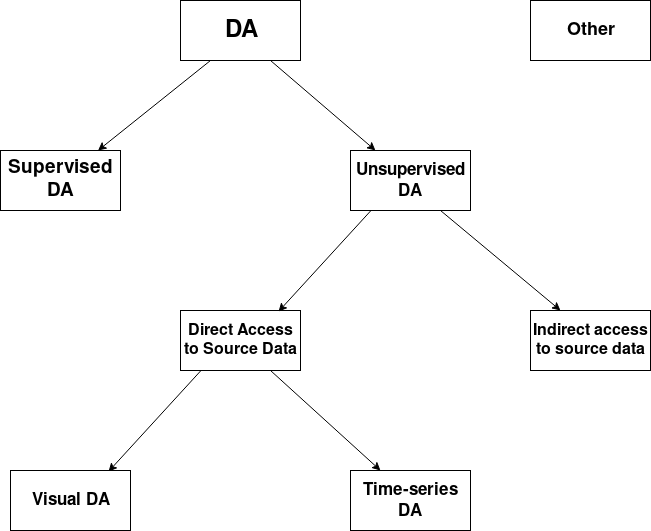}
 \caption{Classification of presented related works in supervised or unsupervised DA. }
 \label{fig:RelatedWork}
\end{figure}

There is a large body of work in DA, mainly focused on the visual domain, which is to a large extent covered by existing surveys  \cite{csurka2017domain,wang2018deep}.  In one of these surveys \cite{wang2018deep} Wang et al. separate domain adaptation based on the type of learning (supervised, semi-supervised, unsupervised), whether or not the feature space is the same in source and target domains, and whether the  work performs  one-step or multi-step DA.  We follow a similar separation, however due to space limitations we focus on the works that are mostly related to ours. Thus, we separate on the basis of supervised or unsupervised DA, with more emphasis on unsupervised DA, and for the unsupervised DA case whether the proposed work has direct or indirect access to the source data (see Figure \ref{fig:RelatedWork}). Furthermore, we  separate for visual and time-series applications or methods. For  the works which resemble our work, we compare and discuss about  specific similarities and differences.

For the case of unsupervised visual DA,  a large body of literature use variations of adversarial domain adaptation for visual applications \cite{long2018conditional,Tzeng_2017_CVPR,ganin2016domain,liu2017unsupervised,long2017deep,liu2016coupled}. A good example of an advanced technique that also includes adversarial elements is CyCADA \cite{hoffman2017cycada}. It is a  technique for unsupervised DA which uses  a team of models trained on a unified loss which consists of task losses, cycle-consistent loss \cite{Zhu_2017_ICCV}, GAN losses and semantic consistency loss. The goal is to adapt representations at pixel and feature level while enforcing cycle and semantic consistency loss.  Other works for visual applications that do not directly include adversarial training include \cite{bousmalis2016domain,Haeusser_2017_ICCV}. Additionally,  works exist which, like our proposed method, make use of pseudo-labelling on the target domain \cite{Yan_2017_CVPR,zhang2015deep,saito2017asymmetric,NIPS2016_6110}. However, all of these approaches take advantage of both source and target domain data during their training phase. Furthermore, they use different methods for classifier adaptation on the conditional distribution of the target domain during their training phase, like entropy minimization, minimization of the conditional distribution's MMD between source and target, pseudo-labelling with multiple classifiers  etc. 

As mentioned,  most of the above techniques   perform unsupervised DA with access to labeled source data and unlabeled target data. However, we focus on DA at the end-user with access  only  to unlabeled target's data and a trained classifier at the source data and labels, making it an inherently harder problem (see  Figure \ref{fig:RelatedWork} Unsupervised DA right leaf). Despite  these additional  restrictions, $SICO$ yields an on-average comparable or superior performance relative to several other well-established unsupervised DA works  for the digit datasets (for example the results reported in \cite{ganin2016domain,tzeng2017adversarial}).  Additional works that  provide a solution to unsupervised DA without direct access to the labeled source data include \cite{wulfmeier2018incremental, li2016revisiting}. Wulfmeier et al. \cite{wulfmeier2018incremental} address the problem of degraded model performance due to continuously shifting environment conditions. They develop incremental adversarial DA with which they redefine DA as a stream of incrementally changing domains, to enable a classifier to adapt for example to changing weather conditions, day night circle, etc. Additionally, they propose an extension to use only target data assuming an additional GAN generator which has learned the encoded feature marginal  distribution of the source data.  Similarly, our method  uses   the source data indirectly via $h_{src}$.  However, we investigate the normal domain adaptation scenario, and not a case of  transitioning incremental domain shifts. Li et al. \cite{li2016revisiting} introduce Adaptive Batch Normalization (AdaBN),  and use only a classifier and the target data. They standardize each neuron's output  with the average  and standard deviation of its output from the images in the target domain. This standardization ensures that each neuron receives data from a similar distribution, no matter of which domain the data come from.  Using this idea, Zhang et al. \cite{zhang2017new}, train a convolutional network with layers  of various kernel sizes for fault diagnosis on raw vibration signals. They then perform AdaBN for the CNN model trained in the source domain. However,  this approach does not   train  with the target domain data, and is only  a form of cross domain normalization for the trained classifier. As a result  we hypothesize that it does not take full advantage of the  knowledge potential of the target sample. 

For the case of unsupervised DA for time-series data,  Purushotham et al.  \cite{purushotham2016variational}, use  Variational Reccurent Neural Nets \cite{chung2015recurrent}, and employ adversarial DA to train  in order to achieve DA from the  latent representations. In \cite{aswolinskiy2017unsupervised},  Aswolinskiy et al., propose Unsupervised Transfer Learning via Self-Predictive Modelling. In the proposed approach,  a linear transformation \textit{Q}  is learned that minimizes the error identified by a self-predictive model between the source and the transformed by \textit{Q} target domain. Then a classifier trained in the source domain is applied in the transformed target domain. We assigned this work under the umbrella of unsupervised domain adaptation since, as mentioned in the paper, the authors  present  a  general domain adaptation framework and  because it fits the data access criteria we specify. Other works include \cite{chai2017fast,natarajan2016domain,zhang2017new}.   Chai et al. \cite{chai2017fast} perform EEG domain adaptation for sentiment recognition by using a linear transformation function that matches the marginal distributions of the source and target feature spaces. They additionally employ a pseudo-labelling approach to transfer confident target data to the transformed source domain which relates to our own approach for choosing confident samples. Natarajaan et al. \cite{natarajan2016domain} use DA in order to mitigate the negative effects from the lab-to-field transition for cocaine detection using wearable ECG.  

In the context of supervised DA,  Persello et al. \cite{persello2012active,persello2012interactive},  use Active Learning for DA for classification of remote sensing images. This  approach loosely relates to ours, in the sense that it iteratively trains the classifier with a differentiating training set. However, the proposed approach assumes access to the source data (for data points to be removed from the training set), and a user in the role of supervisor, which assigns labels for  chosen samples.

Finally, other works that are related in the context of our work include  \cite{fawaz2018transfer,zou2018unsupervised,zhang2017curriculum}.  Fawaz et al. \cite{fawaz2018transfer}  use transfer learning for time series classification. They evaluate on the UCR archive for all possible combinations, and use a  method that relies on Dynamic Time Warping to measure the  similarity between the different datasets. Regarding CL, both works included apply DA in the context of semantic segmentation. Zou et al. \cite{zou2018unsupervised} utilize pseudolabels and self-training to learn the new domain. Contrary to our work, they use a loss which contains terms for the loss of the source and the target domain plus a $L_1$-regularization term on the pseudolabels. They optimize in a two-step fashion, optimizing first the pseudolabels and then the classifier weights.
Zhang et al. \cite{zhang2017curriculum} derive their curriculum by separating between the hard task of semantic segmentation from the easy task of learning high-level properties of the unknown labels.
They minimize a joint loss which includes terms from both domains. For their easy task  they infer the target labels and landmark pixel labels by utilizing training and labelling in the source domain (e.g, retrieving the  nearest source neighbors for a target image and  transfer the labelling).


\section{Conclusion}

  The primary motivation for our work is to enable end-users to create  personal classifiers, e.g., for health applications  without labeled data. In particular, we foresee a collaboration in which a  host releases a classifier  $h_{src}$, and  the end-user (a patient in our case),  uses her data and the classifier  to create a new personalized classifier (i.e., adapted to the domain of the end-user). 
 In this work we achieve this  by performing $SICO$. $SICO$, iteratively adapts classifiers to the new domain based on high-confidence data from the previous classifiers, and without the need of the source data and labels. Based on our scenario, we are more interested in the case of performing  DA at the end-user, but obviously $SICO$ can also be  applied at the host.  We apply $SICO$ for the case of sleep apnea detection, and use a large real-world clinical dataset for its evaluation. Additionally, we experiment with two open sleep monitoring datasets and the MNIST, SVHN and USPS datasets for digit classification. By this, we achieve (1) reproducibility,   (2) demonstrate  the generalizability of $SICO$ for another task, and (3) achieve comparability with related work. The results from these experiments show consistently  better performance of $h_{tg}$ in comparison to  $h_{src}$,  as expected.  Depending on the case, we  get an increase in kappa of up to 0.24. For the task of digit classification DA, $SICO$ achieves again a consistently good performance. Despite  the additional limitations of the scenario that we investigate, the performance of several well established related works e.g., ADDA and gradient reversal  is lower than the presented results of this paper.

As a next step, we are interested in investigating how well can the technique be applied if the source classifier is trained under differential privacy guarantees. Another interesting  application we investigate is the combination of the proposed approach with a style transfer method with the goal of increasing the DA capabilities of the approach for tasks containing different types of sensors.



\section*{Acknowledgements}

 This work was performed as part of the CESAR project (nr.
250239/O70) funded by The Research Council of Norway.

 
\bibliographystyle{unsrt}
\bibliography{MyBib}

\begin{thebibliography}{10}

\bibitem{pan2009survey}
Sinno~Jialin Pan and Qiang Yang.
\newblock A survey on transfer learning.
\newblock {\em IEEE Transactions on knowledge and data engineering},
  22(10):1345--1359, 2009.

\bibitem{ganin2014unsupervised}
Yaroslav Ganin and Victor Lempitsky.
\newblock Unsupervised domain adaptation by backpropagation.
\newblock {\em arXiv preprint arXiv:1409.7495}, 2014.

\bibitem{kang2019contrastive}
Guoliang Kang, Lu~Jiang, Yi~Yang, and Alexander~G Hauptmann.
\newblock Contrastive adaptation network for unsupervised domain adaptation.
\newblock In {\em Proceedings of the IEEE Conference on Computer Vision and
  Pattern Recognition}, pages 4893--4902, 2019.

\bibitem{kristiansen2018data}
Stein Kristiansen, Mari~S{\o}nsteby Hugaas, Vera Goebel, Thomas Plagemann,
  Konstantinos Nikolaidis, and Knut Liest{\o}l.
\newblock Data mining for patient friendly apnea detection.
\newblock {\em IEEE Access}, 6:74598--74615, 2018.

\bibitem{nikolaidis2019augmenting}
Konstantinos Nikolaidis, Stein Kristiansen, Vera Goebel, Thomas Plagemann, Knut
  Liest{\o}l, and Mohan Kankanhalli.
\newblock Augmenting physiological time series data: A case study for sleep
  apnea detection.
\newblock {\em arXiv preprint arXiv:1905.09068}, 2019.

\bibitem{xie2012real}
Baile Xie and Hlaing Minn.
\newblock Real-time sleep apnea detection by classifier combination.
\newblock {\em IEEE Transactions on information technology in biomedicine},
  16(3):469--477, 2012.

\bibitem{Fredrikson:2015:MIA:2810103.2813677}
Matt Fredrikson, Somesh Jha, and Thomas Ristenpart.
\newblock Model inversion attacks that exploit confidence information and basic
  countermeasures.
\newblock In {\em Proceedings of the 22Nd ACM SIGSAC Conference on Computer and
  Communications Security}, CCS '15, pages 1322--1333, New York, NY, USA, 2015.
  ACM.

\bibitem{ben2007analysis}
Shai Ben-David, John Blitzer, Koby Crammer, and Fernando Pereira.
\newblock Analysis of representations for domain adaptation.
\newblock In {\em Advances in neural information processing systems}, pages
  137--144, 2007.

\bibitem{bengio2009curriculum}
Yoshua Bengio, J{\'e}r{\^o}me Louradour, Ronan Collobert, and Jason Weston.
\newblock Curriculum learning.
\newblock In {\em Proceedings of the 26th annual international conference on
  machine learning}, pages 41--48, 2009.

\bibitem{hacohen2019power}
Guy Hacohen and Daphna Weinshall.
\newblock On the power of curriculum learning in training deep networks.
\newblock {\em arXiv preprint arXiv:1904.03626}, 2019.

\bibitem{lecun1998gradient}
Yann LeCun, L{\'e}on Bottou, Yoshua Bengio, and Patrick Haffner.
\newblock Gradient-based learning applied to document recognition.
\newblock {\em Proceedings of the IEEE}, 86(11):2278--2324, 1998.

\bibitem{hull1994database}
Jonathan~J. Hull.
\newblock A database for handwritten text recognition research.
\newblock {\em IEEE Transactions on pattern analysis and machine intelligence},
  16(5):550--554, 1994.

\bibitem{netzer2011reading}
Yuval Netzer, Tao Wang, Adam Coates, Alessandro Bissacco, Bo~Wu, and Andrew~Y
  Ng.
\newblock Reading digits in natural images with unsupervised feature learning.
\newblock 2011.

\bibitem{penzel2000apnea}
Thomas Penzel, George~B Moody, Roger~G Mark, Ary~L Goldberger, and J~Hermann
  Peter.
\newblock The apnea-ecg database.
\newblock In {\em Computers in Cardiology 2000. Vol. 27 (Cat. 00CH37163)},
  pages 255--258. IEEE, 2000.

\bibitem{ichimaru1999development}
Yuhei Ichimaru and GB~Moody.
\newblock Development of the polysomnographic database on cd-rom.
\newblock {\em Psychiatry and clinical neurosciences}, 53(2):175--177, 1999.

\bibitem{traaen2019treatment}
Gunn~Marit Traaen, Lars Aaker{\o}y, et~al.
\newblock Treatment of sleep apnea in patients with paroxysmal atrial
  fibrillation: Design and rationale of a randomized controlled trial.
\newblock {\em Scandinavian Cardiovascular Journal}, (52:6,pp. 372-377):1--20,
  January 2019.

\bibitem{cohen1960coefficient}
Jacob Cohen.
\newblock A coefficient of agreement for nominal scales.
\newblock {\em Educational and psychological measurement}, 20(1):37--46, 1960.

\bibitem{kingma2014adam}
Diederik~P Kingma and Jimmy Ba.
\newblock Adam: A method for stochastic optimization.
\newblock {\em arXiv preprint arXiv:1412.6980}, 2014.

\bibitem{csurka2017domain}
Gabriela Csurka.
\newblock Domain adaptation for visual applications: A comprehensive survey.
\newblock {\em arXiv preprint arXiv:1702.05374}, 2017.

\bibitem{wang2018deep}
Mei Wang and Weihong Deng.
\newblock Deep visual domain adaptation: A survey.
\newblock {\em Neurocomputing}, 312:135--153, 2018.

\bibitem{long2018conditional}
Mingsheng Long, Zhangjie Cao, Jianmin Wang, and Michael~I Jordan.
\newblock Conditional adversarial domain adaptation.
\newblock In {\em Advances in Neural Information Processing Systems}, pages
  1640--1650, 2018.

\bibitem{Tzeng_2017_CVPR}
Eric Tzeng, Judy Hoffman, Kate Saenko, and Trevor Darrell.
\newblock Adversarial discriminative domain adaptation.
\newblock In {\em The IEEE Conference on Computer Vision and Pattern
  Recognition (CVPR)}, July 2017.

\bibitem{ganin2016domain}
Yaroslav Ganin, Evgeniya Ustinova, Hana Ajakan, Pascal Germain, Hugo
  Larochelle, Fran{\c{c}}ois Laviolette, Mario Marchand, and Victor Lempitsky.
\newblock Domain-adversarial training of neural networks.
\newblock {\em The Journal of Machine Learning Research}, 17(1):2096--2030,
  2016.

\bibitem{liu2017unsupervised}
Ming-Yu Liu, Thomas Breuel, and Jan Kautz.
\newblock Unsupervised image-to-image translation networks.
\newblock In {\em Advances in neural information processing systems}, pages
  700--708, 2017.

\bibitem{long2017deep}
Mingsheng Long, Han Zhu, Jianmin Wang, and Michael~I Jordan.
\newblock Deep transfer learning with joint adaptation networks.
\newblock In {\em Proceedings of the 34th International Conference on Machine
  Learning-Volume 70}, pages 2208--2217. JMLR. org, 2017.

\bibitem{liu2016coupled}
Ming-Yu Liu and Oncel Tuzel.
\newblock Coupled generative adversarial networks.
\newblock In {\em Advances in neural information processing systems}, pages
  469--477, 2016.

\bibitem{hoffman2017cycada}
Judy Hoffman, Eric Tzeng, Taesung Park, Jun-Yan Zhu, Phillip Isola, Kate
  Saenko, Alexei~A Efros, and Trevor Darrell.
\newblock Cycada: Cycle-consistent adversarial domain adaptation.
\newblock {\em arXiv preprint arXiv:1711.03213}, 2017.

\bibitem{Zhu_2017_ICCV}
Jun-Yan Zhu, Taesung Park, Phillip Isola, and Alexei~A. Efros.
\newblock Unpaired image-to-image translation using cycle-consistent
  adversarial networks.
\newblock In {\em The IEEE International Conference on Computer Vision (ICCV)},
  Oct 2017.

\bibitem{bousmalis2016domain}
Konstantinos Bousmalis, George Trigeorgis, Nathan Silberman, Dilip Krishnan,
  and Dumitru Erhan.
\newblock Domain separation networks.
\newblock In {\em Advances in neural information processing systems}, pages
  343--351, 2016.

\bibitem{Haeusser_2017_ICCV}
Philip Haeusser, Thomas Frerix, Alexander Mordvintsev, and Daniel Cremers.
\newblock Associative domain adaptation.
\newblock In {\em The IEEE International Conference on Computer Vision (ICCV)},
  Oct 2017.

\bibitem{Yan_2017_CVPR}
Hongliang Yan, Yukang Ding, Peihua Li, Qilong Wang, Yong Xu, and Wangmeng Zuo.
\newblock Mind the class weight bias: Weighted maximum mean discrepancy for
  unsupervised domain adaptation.
\newblock In {\em The IEEE Conference on Computer Vision and Pattern
  Recognition (CVPR)}, July 2017.

\bibitem{zhang2015deep}
Xu~Zhang, Felix~Xinnan Yu, Shih-Fu Chang, and Shengjin Wang.
\newblock Deep transfer network: Unsupervised domain adaptation.
\newblock {\em arXiv preprint arXiv:1503.00591}, 2015.

\bibitem{saito2017asymmetric}
Kuniaki Saito, Yoshitaka Ushiku, and Tatsuya Harada.
\newblock Asymmetric tri-training for unsupervised domain adaptation.
\newblock In {\em Proceedings of the 34th International Conference on Machine
  Learning-Volume 70}, pages 2988--2997. JMLR. org, 2017.

\bibitem{NIPS2016_6110}
Mingsheng Long, Han Zhu, Jianmin Wang, and Michael~I Jordan.
\newblock Unsupervised domain adaptation with residual transfer networks.
\newblock In D.~D. Lee, M.~Sugiyama, U.~V. Luxburg, I.~Guyon, and R.~Garnett,
  editors, {\em Advances in Neural Information Processing Systems 29}, pages
  136--144. Curran Associates, Inc., 2016.

\bibitem{tzeng2017adversarial}
Eric Tzeng, Judy Hoffman, Kate Saenko, and Trevor Darrell.
\newblock Adversarial discriminative domain adaptation.
\newblock In {\em Proceedings of the IEEE Conference on Computer Vision and
  Pattern Recognition}, pages 7167--7176, 2017.

\bibitem{wulfmeier2018incremental}
Markus Wulfmeier, Alex Bewley, and Ingmar Posner.
\newblock Incremental adversarial domain adaptation for continually changing
  environments.
\newblock In {\em 2018 IEEE International conference on robotics and automation
  (ICRA)}, pages 1--9. IEEE, 2018.

\bibitem{li2016revisiting}
Yanghao Li, Naiyan Wang, Jianping Shi, Jiaying Liu, and Xiaodi Hou.
\newblock Revisiting batch normalization for practical domain adaptation.
\newblock {\em arXiv preprint arXiv:1603.04779}, 2016.

\bibitem{zhang2017new}
Wei Zhang, Gaoliang Peng, Chuanhao Li, Yuanhang Chen, and Zhujun Zhang.
\newblock A new deep learning model for fault diagnosis with good anti-noise
  and domain adaptation ability on raw vibration signals.
\newblock {\em Sensors}, 17(2):425, 2017.

\bibitem{purushotham2016variational}
Sanjay Purushotham, Wilka Carvalho, Tanachat Nilanon, and Yan Liu.
\newblock Variational recurrent adversarial deep domain adaptation.
\newblock 2016.

\bibitem{chung2015recurrent}
Junyoung Chung, Kyle Kastner, Laurent Dinh, Kratarth Goel, Aaron~C Courville,
  and Yoshua Bengio.
\newblock A recurrent latent variable model for sequential data.
\newblock In {\em Advances in neural information processing systems}, pages
  2980--2988, 2015.

\bibitem{aswolinskiy2017unsupervised}
Witali Aswolinskiy and Barbara Hammer.
\newblock Unsupervised transfer learning for time series via self-predictive
  modelling-first results.
\newblock In {\em Proceedings of the Workshop on New Challenges in Neural
  Computation (NC2)}, volume~3, 2017.

\bibitem{chai2017fast}
Xin Chai, Qisong Wang, Yongping Zhao, Yongqiang Li, Dan Liu, Xin Liu, and
  Ou~Bai.
\newblock A fast, efficient domain adaptation technique for cross-domain
  electroencephalography (eeg)-based emotion recognition.
\newblock {\em Sensors}, 17(5):1014, 2017.

\bibitem{natarajan2016domain}
Annamalai Natarajan, Gustavo Angarita, Edward Gaiser, Robert Malison, Deepak
  Ganesan, and Benjamin~M Marlin.
\newblock Domain adaptation methods for improving lab-to-field generalization
  of cocaine detection using wearable ecg.
\newblock In {\em Proceedings of the 2016 ACM International Joint Conference on
  Pervasive and Ubiquitous Computing}, pages 875--885. ACM, 2016.

\bibitem{persello2012active}
Claudio Persello and Lorenzo Bruzzone.
\newblock Active learning for domain adaptation in the supervised
  classification of remote sensing images.
\newblock {\em IEEE Transactions on Geoscience and Remote Sensing},
  50(11):4468--4483, 2012.

\bibitem{persello2012interactive}
Claudio Persello.
\newblock Interactive domain adaptation for the classification of remote
  sensing images using active learning.
\newblock {\em IEEE geoscience and remote sensing letters}, 10(4):736--740,
  2012.

\bibitem{fawaz2018transfer}
Hassan~Ismail Fawaz, Germain Forestier, Jonathan Weber, Lhassane Idoumghar, and
  Pierre-Alain Muller.
\newblock Transfer learning for time series classification.
\newblock In {\em 2018 IEEE International Conference on Big Data (Big Data)},
  pages 1367--1376. IEEE, 2018.

\bibitem{zou2018unsupervised}
Yang Zou, Zhiding Yu, BVK Vijaya~Kumar, and Jinsong Wang.
\newblock Unsupervised domain adaptation for semantic segmentation via
  class-balanced self-training.
\newblock In {\em Proceedings of the European conference on computer vision
  (ECCV)}, pages 289--305, 2018.

\bibitem{zhang2017curriculum}
Yang Zhang, Philip David, and Boqing Gong.
\newblock Curriculum domain adaptation for semantic segmentation of urban
  scenes.
\newblock In {\em Proceedings of the IEEE International Conference on Computer
  Vision}, pages 2020--2030, 2017.

\end{thebibliography}
\clearpage

\appendix

\section{Additional Design Decisions}

In this Appendix, we discuss additional design decisions made during the algorithm. 
\begin{itemize}
    \item \textbf{Use of hard or soft labels for labelling of} $D_{CLi},i=0..N$: In the original approach  we use as labels the one-hot encoding vector based on the maximum argument of the output class probabilities of the classifier which is doing the labelling for the particular datapoint. Thus each new classifier is trained on one-hot encoding labels. Alternatively, we can utilize directly the output class probabilities to act as labels and for the training of each new classifier. During our experiments, we experimented with this approach, however in almost all cases it yielded worst results than the use of hard labels. As a result we use hard labelling in our final experiments. The only exception is the $MB$ dataset, for which we hypothesize that due to the poor labelling quality, there is a stronger discrepancy between the true conditional distribution, and the hard labelling. This would also result in the optimal (Bayes) classifier  having a very high true risk because of the poor labelling. Thus since we have a finite amount of data, using hard labels could be misrepresentative of the previous  classifiers' class probabilities.  

\item \textbf{Training Iterations for $h_{CLi}$:} Depending on the sizes of the different datasets we train for a minimum of  10 ($A3$) epochs to a maximum of 50 epochs ($MB,AE$).     

\end{itemize}
\section{Accuracy, Specificity, Sensitivity for Apnea Detection Experiments}
Appendix B shows the Accuracy, Sensitivity and Specificity for the Apnea detection experiments DA experiments.

\begin{table}[h]
\caption{}
\centering
\begin{tabular}{ | p{2cm}| p{2cm}|p{2cm}|p{2cm}|}
\hline
\multicolumn{4}{|c|}{ $SICO$ $h_{tg}$  Accuracy for NAF} \\
\hline
NAF: &$h_{src}$,$D_{src}$&$h_{src}$,$D_{tg}$ &$h_{tg},D_{tg}$ \\
\hline
$A3\rightarrow AE$:&84.75$\pm$0.11&84.37$\pm$5.62&92.16$\pm$2.18\\
$A3\rightarrow MB$:&84.75$\pm$0.11&54.72$\pm$0.58&59.56$\pm$0.89\\
 $AE\rightarrow MB$:&97.32$\pm$0.24&55.88$\pm$0.57&56.60$\pm$0.33\\
 $MB\rightarrow AE$:&70.88$\pm$0.79&82.90$\pm$1.54&89.60$\pm$1.13\\
 $AE\rightarrow A3$:&97.32$\pm$0.24&64.43$\pm$0.64&68.71$\pm$1.00\\
\hline
\end{tabular}
\begin{tabular}{ | p{2cm}| p{2cm}|p{2cm}|p{2cm}|}
\hline
\multicolumn{4}{|c|}{ $SICO$ $h_{tg}$  Sensitivity for NAF} \\
\hline
NAF: &$h_{src}$,$D_{src}$&$h_{src}$,$D_{tg}$ &$h_{tg},D_{tg}$ \\
\hline
$A3\rightarrow AE$:&85.82$\pm$0.78&77.82$\pm$0.50&94.99$\pm$1.54\\
$A3\rightarrow MB$:&85.82$\pm$0.66&26.29$\pm$2.48&52.12$\pm$3.54\\
 $AE\rightarrow MB$:&96.08$\pm$0.24&48.50$\pm$0.74&58.34$\pm$1.77\\
 $MB\rightarrow AE$:&73.38$\pm$0.87&86.66$\pm$2,23&95.19$\pm$0.76\\
 $AE\rightarrow A3$:&96.08$\pm$0.24&56.78$\pm$2.32&74.32$\pm$0.96\\
\hline
\end{tabular}

\begin{tabular}{ | p{2cm}| p{2cm}|p{2cm}|p{2cm}|}
\hline
\multicolumn{4}{|c|}{ $SICO$ $h_{tg}$  Specificity for NAF} \\
\hline
NAF: &$h_{src}$,$D_{src}$&$h_{src}$,$D_{tg}$ &$h_{tg},D_{tg}$ \\
\hline
$A3\rightarrow AE$:&83.50$\pm$0.73&88.93$\pm$4.59&90.29$\pm$4.24\\
$A3\rightarrow MB$:&83.50$\pm$0.73&84.06$\pm$1.95&67.23$\pm$3.71\\
 $AE\rightarrow MB$:&98.13$\pm$0.31&63.49$\pm$1.16&54.79$\pm$1.37\\
 $MB\rightarrow AE$:&68.28$\pm$1.76&80.42$\pm$2.29&85.92$\pm$1.55\\
 $AE\rightarrow A3$:&98.13$\pm$0.31&73.39$\pm$1.50&62.14$\pm$1.21\\
\hline
\end{tabular}

\label{table:ApneaAcc}
\end{table}

\begin{table}[h]
\centering
\caption{}
\begin{tabular}{ | p{3.5cm}| p{2cm}|p{2cm}|p{2cm}|}
\hline
\multicolumn{4}{|c|}{  $SICO$ $h_{tg}$ Accuracy for different sensors} \\
\hline
 &Resp A &Resp C&Sp02\\
\hline
$A3 \rightarrow AE$: $h_{src}$,$D_{src}$:&83.51$\pm$1.92&83.57$\pm$0.16&85.97$\pm$0.30\\
$A3 \rightarrow AE$: $h_{src}$,$D_{tg}$: &89.74$\pm$0.81&77.68$\pm$4.89&80.12$\pm$2.4\\
$A3 \rightarrow AE$: $h_{tg},D_{tg}$:&90.42$\pm$0.43&90.88$\pm$0.56&87.44$\pm$1.01\\
\hline
$AE \rightarrow A3$: $h_{src}$,$D_{src}$:&96.32$\pm$0.24&95.66$\pm$0.02&94.76$\pm$0.17\\
$AE \rightarrow A3$: $h_{src}$,$D_{tg}$: &61.79$\pm$0.59&59.49$\pm$1.08&47.32$\pm$1.26\\
$AE \rightarrow A3$: $h_{tg},D_{tg}$:&74.44$\pm$0.58&63.20$\pm$0.23&60.34$\pm$0.78\\
\hline
\end{tabular}

\begin{tabular}{ | p{3.5cm}| p{2cm}|p{2cm}|p{2cm}|}
\hline
\multicolumn{4}{|c|}{  $SICO$ $h_{tg}$ Sensitivity for different sensors} \\
\hline
 &Resp A &Resp C&Sp02\\
\hline
$A3 \rightarrow AE$: $h_{src}$,$D_{src}$:&86.3$\pm$0.51&85.22$\pm$0.66&85.40$\pm$1.05\\
$A3 \rightarrow AE$: $h_{src}$,$D_{tg}$: &93.83$\pm$5.15&92.16$\pm$4.44&94.64$\pm$0.53\\
$A3 \rightarrow AE$: $h_{tg},D_{tg}$:&98.48$\pm$0.07&96.91$\pm$0.75&89.44$\pm$0.79\\
\hline
$AE \rightarrow A3$: $h_{src}$,$D_{src}$:&95.85$\pm$0.43&97.46$\pm$0.52&97.48$\pm$0.13\\
$AE \rightarrow A3$: $h_{src}$,$D_{tg}$: &30.76$\pm$1.27&27.92$\pm$2.39&9.53$\pm$1.29\\
$AE \rightarrow A3$: $h_{tg},D_{tg}$:&77.46$\pm$0.54&65.51$\pm$0.70&69.16$\pm$0.97\\
\hline
\end{tabular}

\begin{tabular}{ | p{3.5cm}| p{2cm}|p{2cm}|p{2cm}|}
\hline
\multicolumn{4}{|c|}{  $SICO$ $h_{tg}$ Specificity for different sensors} \\
\hline
 &Resp A &Resp C&Sp02\\
\hline
$A3 \rightarrow AE$: $h_{src}$,$D_{src}$:&80.55$\pm$0.60&81.63$\pm$0.46&86.63$\pm$1.86\\
$A3 \rightarrow AE$: $h_{src}$,$D_{tg}$: &87.04$\pm$2.56&68.17$\pm$9.66&70.59$\pm$3.69\\
$A3 \rightarrow AE$: $h_{tg},D_{tg}$:&85.12$\pm$.0.73&86.91$\pm$0.91&86.12$\pm$1.53\\
\hline
$AE \rightarrow A3$: $h_{src}$,$D_{src}$:&96.61$\pm$0.24&94.43$\pm$0.46&97.48$\pm$0.13\\
$AE \rightarrow A3$: $h_{src}$,$D_{tg}$: &98.17$\pm$0.20&96.49$\pm$0.48&90.40$\pm$1.16\\
$AE \rightarrow A3$: $h_{tg},D_{tg}$:&70.91$\pm$0.87&60.50$\pm$0.54&50.00$\pm$1.06\\
\hline
\end{tabular}

 \label{table:ApneaOther}
\end{table}

A characteristic which was not identifiable from  the kappa values is the positive balancing effect that $SICO$ has between specificity and sensitivity. When using $h_{src}$,  notice that in many cases we obtain high specificity at the expense of high sensitivity (e.g., NAF:$A3\rightarrow MB$, NAF:$AE\rightarrow MB$). We observe that for $h_{tg}$ this effect is minimized. This  means that $h_{tg}$ obtains increased sensitivity  compared  to $h_{src}$. This is a  positive characteristic since sensitivity (i.e., the percentage of  the correctly detected apneic minutes) plays a  crucial role in the context of sleep apnea detection as low sensitivity can lead to a falsely negative diagnosis, making the system inherently untrustworthy.

\end{document}